\documentclass[journal]{IEEEtran}
\usepackage{cite}
\usepackage{amsmath}
\usepackage{amssymb}
\usepackage{bm}
\usepackage{multicol}
\usepackage{multirow}
\usepackage{booktabs}
\usepackage{tabularx}
\usepackage{color}
\usepackage[pdftex]{graphicx}
\usepackage[setpagesize=false]{hyperref}
\usepackage{url}

\begin{document}
%
\title{Deep Neural Generative Model of Functional MRI Images for Psychiatric Disorder Diagnosis}
%
%
%

\author{Takashi~Matsubara,~\IEEEmembership{Member,~IEEE,}
        Tetsuo~Tashiro,~\IEEEmembership{Nonmember,}
        and~Kuniaki~Uehara,~\IEEEmembership{Nonmember}
\thanks{T. Matsubara, T. Tashiro, and K. Uehara are with the Graduate School of System Informatics, Kobe University, Hyogo, Japan e-mail: matsubara@phoenix.kobe-u.ac.jp.}
\thanks{Manuscript received April 19, 2005; revised August 26, 2015.}}

%
%

\markboth{Journal of \LaTeX\ Class Files,~Vol.~14, No.~8, August~2015}%
{Matsubara \MakeLowercase{\textit{et al.}}: Deep Neural Generative Model of Functional MRI Images for Psychiatric Disorder Diagnosis}
%



\maketitle

\begin{abstract}
Accurate diagnosis of psychiatric disorders plays a critical role in improving the quality of life for patients and potentially supports the development of new treatments.
Many studies have been conducted on machine learning techniques that seek brain imaging data for specific biomarkers of disorders.
These studies have encountered the following dilemma: A direct classification overfits to a small number of high-dimensional samples but unsupervised feature-extraction has the risk of extracting a signal of no interest.
In addition, such studies often provided only diagnoses for patients without presenting the reasons for these diagnoses.
This study proposed a deep neural generative model of resting-state functional magnetic resonance imaging (fMRI) data.
The proposed model is conditioned by the assumption of the subject's state and estimates the posterior probability of the subject's state given the imaging data, using Bayes' rule.
This study applied the proposed model to diagnose schizophrenia and bipolar disorders.
Diagnostic accuracy was improved by a large margin over competitive approaches, namely classifications of functional connectivity, discriminative/generative models of region-wise signals, and those with unsupervised feature-extractors.
The proposed model visualizes brain regions largely related to the disorders, thus motivating further biological investigation.
\end{abstract}

\begin{IEEEkeywords}
deep learning, generative model, functional magnetic resonance imaging, psychiatric-disorder diagnosis, schizophrenia, bipolar disorder
\end{IEEEkeywords}

%
\IEEEpeerreviewmaketitle

\section{Introduction}
\IEEEPARstart{A}{ccurate} diagnosis of neurological and psychiatric disorders plays a critical role in improving quality of life for patients; it provides an opportunity for appropriate treatment and prevention of further disease progression.
Moreover, it potentially enables the effectiveness of treatments to be evaluated and supports the development of new treatments.
With advances in brain imaging techniques such as (functional) magnetic resonance imaging (MRI) and positron emission tomography (PET)~\cite{Sejnowski2014}, many studies have attempted to find specific biomarkers of neurological and psychiatric disorders in brain images using machine learning techniques~\cite{Atkinson2001}, e.g., for schizophrenia~\cite{Castro2016,Plis2014}, Alzheimer's disease (AD)~\cite{Suk2016,Suk2017}, and others~\cite{Zhang2005a,Vergun2016,Abraham2017,Guo2017}.
Resting-state fMRI (rs-fMRI) has received considerable attention~\cite{Zhang2005a,Plis2014,Suk2016,Vergun2016,Abraham2017,Suk2017,Guo2017}.
This approach visualizes interactions among brain regions in subjects at rest, that is, it does not require subjects to perform tasks and to receive stimuli, which eliminates potential confounders, e.g., individual task-skills~\cite{Biswal1995}.

Although neuroimaging datasets continue to increase in size~\cite{Sejnowski2014}, each dataset contains only a small number of high-dimensional samples compared to datasets for other machine-learning tasks.
Unsophisticated application of machine-learning techniques tends to overfit to training samples and to fail in generalizing to unknown samples.
Many existing techniques employed Pearson correlation coefficients (PCC) as a feature, whereby the PCCs were considered to represent the functional connectivity between brain regions~\cite{Castellanos2013,Watanabe2017,Guo2017}.
Then, the techniques consist of feature-selection, dimension-reduction, and classification.
Instead of the PCCs, other studies employed unsupervised dimension-reduction such as principal components analysis (PCA) and independent components analysis (ICA)~\cite{Zhang2005a,Plis2014,Vergun2016,Suk2016} in order to identify low-dimensional dominant patterns directly in each frame or each time-window and to extract the former as features.
Then, these studies diagnosed subjects using supervised classifiers.
These unsupervised feature-selection and dimension-reduction approaches are considered to reduce the risk of overfitting.
However, they inevitably risk extracting factors unrelated to the disorder, rather than extracting disorder-related brain activity~\cite{Lin2016}.

In contrast, artificial neural networks with deep architectures (\emph{deep neural networks}; DNNs) are attracting attention in the machine-learning field (see \cite{Bengio2012,Schmidhuber2015} for a review).
They have the ability to approximate arbitrary functions and learn high-level features from a given dataset automatically, and thereby improve performance in classification and regression tasks related to images, speech, natural language, and more besides.
Variations of DNNs have been employed for neuroimaging datasets.
A multilayer perceptron (MLP) has been employed as a supervised classifier~\cite{Castro2016,Vergun2016}.
An autoencoder (AE) and its variations such as variational autoencoder (VAE)~\cite{Kingma2014} and adversarial autoencoder (AAE)~\cite{Makhzani2015} also have been employed as an unsupervised feature-extractor~\cite{Suk2016,Guo2017}.
These approaches share common difficulties with the aforementioned techniques but they are uniquely characterized by their modifiable structures: The AE can be extended to a \emph{deep neural generative model} (DGM), which implements relationships between multiple factors (e.g., fMRI images, class labels, imposed tasks, and stimuli) in its network structure~\cite{Kingma2014,Kingma2014a,Sohn2015,Maaloe2015,Makhzani2015}.
The DGM with class labels is no longer just an unsupervised feature-extractor but is a generative model of the joint distribution of data points and class labels.
Using Bayes' rule, the DGM also works as a supervised classifier~\cite{Lasserre2006,Prasad2017,Matsubara2017StockPred}.
Hence, the DGM has the aspects of both a supervised classifier and an unsupervised feature-extractor.
Several studies have compared simple discriminative and generative models (i.e., logistic regression and naive Bayes).
They have revealed theoretically and experimentally that the generative model classifies a small-sized dataset better than the discriminative model~\cite{Ng2001,Lasserre2006,Prasad2017}.
While this relationship is not guaranteed to hold for more complicated models like deep neural networks, a DGM potentially overcomes the difficulties that both conventional supervised classifiers and unsupervised feature-extractors encounter.

Given the above, this paper proposes a machine-learning-based method of diagnosing psychiatric disorders using a DGM of rs-fMRI images.
Our proposed DGM considers three factors: a feature obtained from an fMRI image, a class label (controls or patients), and the remaining frame-wise variability.
The frame-wise variability is assumed to represent temporal states of dynamic functional connectivity, what a subject has in mind at that moment, and other factors that vary over time.
It also contains signal of no interest (e.g., body motion that preprocessing does not remove successfully).
Each subject is expected to belong to one of the classes.
Each scan image obtained from a subject is considered to be generated given the subject's class and the remaining frame-wise variability.
Then, if a subject's images are more likely generated given the class of patients rather than the class of controls, the subject is considered to have the disorder because of Bayes' rule.
Since our proposed DGM explicitly has the class label as a visible variable, unlike the ordinary AE, it is free from the risk of not extracting activity of interest.
Furthermore, we propose a method for the proposed DGM to evaluate the contribution weight of each brain region to the diagnosis, which potentially provides a score that assesses the disorder progression.

We evaluate our proposed DGM using open rs-fMRI datasets of schizophrenia and bipolar disorders provided by OpenfMRI (\url{https://openfmri.org/dataset/ds000030/}).
We obtained a region-wise feature vector from each fMRI image by using an automated anatomical labeling (AAL) template \cite{Tzourio-Mazoyer2002}.
Our experimental results demonstrate that our proposed DGM achieves better diagnostic accuracy than existing PCC-based approaches~\cite{Yahata2016,Shen2010}, frame-wise classification using MLP, Gaussian mixture model (GMM), and MLP with AE,~\cite{Zhang2005a,Plis2014,Vergun2016}, and models of temporal dynamics such as hidden Markov model (HMM)~\cite{Suk2016,Vidaurre2017} and long short-term memory (LSTM)~\cite{Dvornek2017}.
Comparisons between generated ROI-wise feature vectors under the assumption of controls or patients visualize regions that contribute to an accurate diagnosis.
A preliminary and limited result of this model may be found in symposium proceedings~\cite{Tashiro2017NOLTA}.

\subsection*{Novelty and Significance}
Before we end the Introduction, we summarize again the main novelty of our proposed DGM compared to other approaches:
\begin{itemize}
  \item Existing studies of fMRI image analysis have employed DGMs as feature extractors~\cite{Suk2016,Guo2017}; they inevitably have a risk of extracting factors unrelated to the disorder instead of disorder-related brain activity~\cite{Lin2016}.
  The proposed DGM works as a classifier directly and is free from such a risk.
  \item Typical DGMs deal with individual samples obtained from a dataset, such as a hand-written digit dataset~\cite{Kingma2014,Kingma2014a,Maaloe2015,Makhzani2015}.
  In Section~\ref{sec:generativemodel}, we derive a generative model of an image set obtained from a subject, which is applicable to other biomedical datasets such as electroencephalogram (EEG) data.
  \item For (semi-)supervised classification, extant studies employed DGMs with discriminative models (feedforward MLPs) $q(y|x)$ as internal components~\cite{Kingma2014a,Maaloe2015,Makhzani2015}; they still have a larger risk of overfitting of the discriminative models $q(y|x)$.
  In contrast, our proposed DGM does not have such a discriminative model $q(y|x)$ but employs Bayes' rule for classification; it works well for a small-sized dataset~\cite{Lasserre2006,Prasad2017,Matsubara2017StockPred} and achieves higher diagnostic accuracies as shown in Section~\ref{sec:accuracy}.
  \item Unlike MLP and classifiers with feature extractions, the proposed DGM can measure the contribution weight of each brain region to the diagnosis as shown in Section~\ref{sec:contribution}.
  This potentially evaluates the disorder progression and contributes to further biomedical investigations of the underlying mechanisms.
\end{itemize}

\section{Deep Neural Generative Model}\label{sec:methods}
\subsection{Generative Model of FMRI Images}\label{sec:generativemodel}
\begin{figure}[t]\centering
  \includegraphics[scale=0.5]{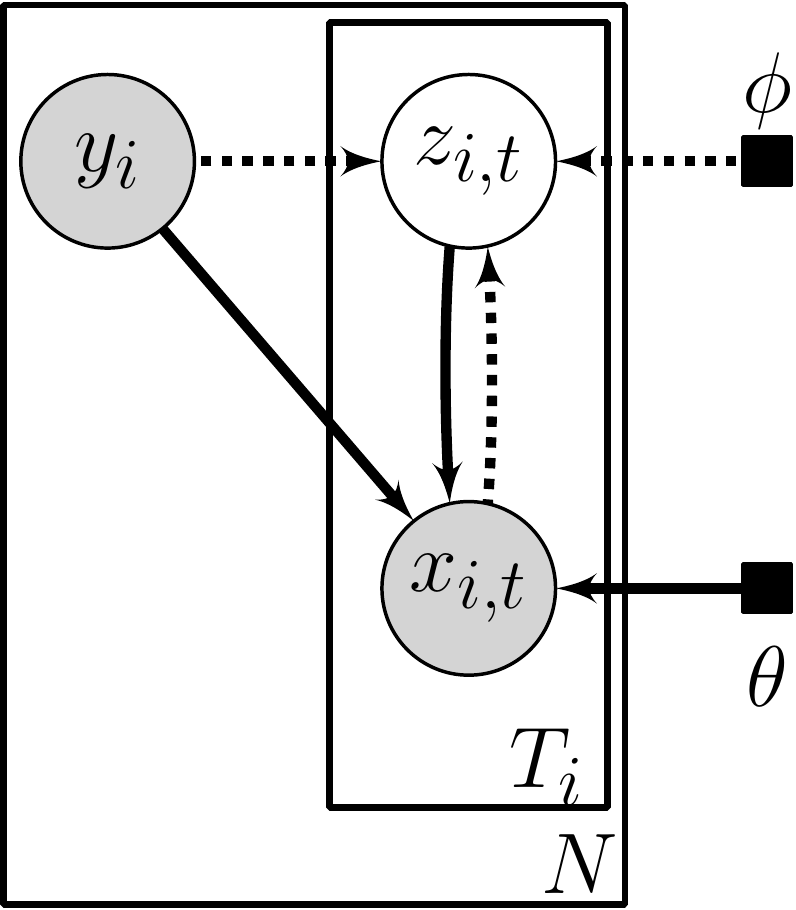}
    \caption{Our proposed generative model of fMRI features (fMRI images or extracted feature vectors) $x_{i,t}$ with diagnosis $y_i$ and remaining variabilities $z_{i,t}$.}\label{fig:gmodel}
\end{figure}

In this section, we propose a generative model of a dataset $\mathcal D$ of fMRI features (fMRI images or extracted feature vectors) and diagnoses.
The dataset $\mathcal D$ contains $N$ subjects indexed by $i$.
Each subject $i$ belongs to a class $y_i$, which is typically represented by a binary value: control $y_i=0$ or patient $y_i=1$.
Each subject $i$ is scanned for $T_i$ frames, providing a subject-wise set $\bm x_i=\{x_{i,t}\}_{t=1}^{T_i}$ of fMRI features $x_{i,t}$.
Then, the complete dataset consists of pairs of all the fMRI features $\bm X=\{\bm x_i\}_{i=1}^{N}$ and the class labels $\bm y=\{y_i\}_{i=1}^{N}$ of subjects $i=1,\dots,N$.

We assume each fMRI feature $x_{i,t}$ is associated with an unobservable latent variable $z_{i,t}$ as well as the subject's class $y_i$.
The latent variable $z_{i,t}$ is not related to the class label $y_i$ but represents frame-wise variability, e.g., brain activity related to subject's cognition at that moment, body motion not removed successfully by preprocessing, and so on.
For simplicity, we employ a time-invariant prior $p(z)$ of the frame-wise variability $z_{i,t}$.
Given the above, we build a frame-wise conditional generative model $p_\theta$ of fMRI features $x_{i,t}$ parameterized by $\theta$.
This is depicted in Fig.~\ref{fig:gmodel} and expressed as
\begin{align*}
  \begin{split}
  p_\theta(x_{i,t}|y_i)
  &=\displaystyle \int_{z_{i,t}} p_\theta(x_{i,t},z_{i,t}|y_i)\\
  &=\displaystyle \int_{z_{i,t}} p_\theta(x_{i,t}|z_{i,t},y_i)p(z_{i,t}).\\
  \end{split}
\end{align*}

Although the posterior $p_\theta(z_{i,t}|x_{i,t},y_i)$ of the latent variable $z_{i,t}$ is required to train the above model $p_\theta$, it is typically intractable.
Based on the variational method~\cite{Blei2017}, the model evidence $\log p_\theta(x_{i,t}|y_i)$ is bounded using an inference model $q_\phi(z_{i,t}|x_{i,t},y_i)$ parameterized by $\phi$ as
\begin{align}
  \begin{split}
  \log p_\theta(x_{i,t}|y_i)
  &=\displaystyle \mathbb E_{q_\phi(z_{i,t}|x_{i,t},y_i)}\left[\log\frac{p_\theta(x_{i,t},z_{i,t}|y_i)}{p_\theta(z_{i,t}|x_{i,t},y_i)}\right]\\
  &=\displaystyle \mathbb E_{q_\phi(z_{i,t}|x_{i,t},y_i)}\left[\log\frac{p_\theta(x_{i,t},z_{i,t}|y_i)}{q_\phi(z_{i,t}|x_{i,t},y_i)}\right]\\
  &\ \ \ +D_{KL}(q_\phi(z_{i,t}|x_{i,t},y_i)||p_\theta(z_{i,t}|x_{i,t},y_i))\\
  &\ge\displaystyle \mathbb E_{q_\phi(z_{i,t}|x_{i,t},y_i)}\left[\log\frac{p_\theta(x_{i,t},z_{i,t}|y_i)}{q_\phi(z_{i,t}|x_{i,t},y_i)}\right]\\
  &=-D_{KL}(q_\phi(z_{i,t}|x_{i,t},y_i)||p(z)) \\
  &\ \ \ + \mathbb E_{q_\phi(z_{i,t}|x_{i,t},y_i)}\left[\log p_\theta(x_{i,t}|z_{i,t},y_i)\right]\\
  &=:\mathcal L(x_{i,t};y_i),
  \end{split}\label{eq:scanlikelihood}
\end{align}
where $D_{KL}(\cdot||\cdot)$ is the Kullback-Leibler divergence and $\mathcal L(x_{i,t};y_i)$ is the evidence lower bound.
This model is built for a single fMRI feature and the same as the class conditional variational autoencoder (CVAE)~\cite{Kingma2014a,Sohn2015}.
Based on this model, we build a structured generative model for each subject and for a dataset depicted in Fig.~\ref{fig:gmodel}.

Because the fMRI features $x_{i,t}$ are assumed to be obtained independently from each other, the subject-wise conditional generative model $p_\theta(\bm x_{i,t}|y_i)$ and its evidence lower bound $\mathcal L(\bm x_i;y_i)$ are simply the frame-wise sum:
\begin{align}
  \begin{split}
  \log p_\theta(\bm x_i|y_i)
  &= \sum_{t=1}^{T_i} \log p_\theta(x_{i,t}|y_i)\\
  &\ge\sum_{t=1}^{T_i}\mathcal L(x_{i,t};y_i)\\
  &=:\mathcal L(\bm x_i;y_i).
  \end{split}\label{eq:subjectlikelihood}
\end{align}
Additionally, the conditional generative model $p_\theta(\bm X|\bm y)$ of the complete dataset and its evidence lower bound $\mathcal L(\bm X;\bm y)$ are expressed as the sum of the subject-wise models:
\begin{align}
  \begin{split}
  \log p_\theta(\bm X|\bm y)
  &= \sum_{i=1}^{N} \log p_\theta(\bm x_{i,t}|y_i)\\
  &\ge\sum_{i=1}^{N}\mathcal L(\bm x_{i,t};y_i)\\
  &=:\mathcal L(\bm X;\bm y).
  \end{split}\label{eq:wholelikelihood}
\end{align}
In general, the evidence lower bound $\mathcal L(\bm X;\bm y)$ of the complete dataset is the objective function of the parameters $\theta$ and $\phi$ of the conditional generative model $p_\theta$ and the inference model $q_\phi$ to be maximized.
In practice, we train the frame-wise model $p_\theta(x_{i,t}|y_i)$ to maximize its evidence lower bound $\mathcal L(x_{i,t};y_i)$, and thereby train the conditional generative model $p_\theta(\bm X|\bm y)$ of the complete dataset.

\subsection{Intuitive Comparison with Existing Methods}
We employed the time-invariant prior $p(z)$ and we did not explicitly model temporal dynamics of the frame-wise variability $z_{i,t}$.
While this means an assumption that each frame $x_{i,t}$ is modeled and associated with the class label $y_i$ individually, this does not imply independence of the frame-wise variability $z_{i,t}$ and the fMRI feature $x_{i,t}$ at the different timepoints $t$.
Since we did not impose any constraints on the posterior $p(z_{i,t}|x_{i,t},y)$ of the frame-wise variability $z_{i,t}$ at different timepoints $t$, the posterior $p(z_{i,t}|x_{i,t},y)$ of adjacent scans $x_{i,t}$ are allowed to be similar to each other and to capture a temporal dynamics.
On the other hand, many existing studies explicitly focused on the functional connectivity averaged over time~\cite{Castellanos2013,Watanabe2017,Shen2010,Tang2012,Yahata2016,Takagi2017} or dynamic functional connectivity~\cite{Liu2013,Li2015f,Suk2016} among the regions.
We discuss and clarify why our proposed model works in Sections~\ref{sec:dgmfc} and \ref{sec:reconstruction}.

If the class label $y$ is the sole latent variable, the generative model outputs only two prototypical posterior distributions $p_\theta(x|y)$ depending on the class label $y=0$ or $y=1$; its representational power is strictly limited.
We enriched the representational power of the proposed generative model $p_\theta(x|y)=\int_z p_\theta(x|z,y)p(z)$ by using the additional latent variable $z$.
Thanks to the additional latent variable $z$, the generative model $p_\theta(x|y)$ can be a complicated distribution, which is a mixture of various posterior distributions $p_\theta(x|y,z)$ depending on the latent variable $z$.

Principal component analysis (PCA) and autoencoder (AE) variations have been used as an unsupervised feature-extractor~\cite{Zhang2005a,Plis2014,Vergun2016,Suk2016,Guo2017}.
These models can be expressed as $p(x)=\int_z p(x|z)p(z)$.
In this case, the latent variable $z$ contains high-level relations among the input variables, including disorder-related signals.
Owing to the nature of unsupervised learning, this model has a risk of extracting salient but disorder-unrelated signals instead of a signal of interest.
Conversely, our proposed generative model has the additional class label $y$ as a latent variable; the model assumes that the latent variables $y$ and $z$ are independent of each other and jointly serve as a compressed representation of the visible variable $x$.
Then, we expect that the disorder-related information in the visible variable $x$ is associated only with the class label $y$ but not with the latent variable $z$, unlike in PCA and AE variations.
Since our purpose is diagnosing subjects (i.e., inferring the class label $y$ as a query variable), the latent variable $z$ can be considered as a nuisance variable.
This kind of disentanglements of hidden information between the class label and others has been found in various studies on generative models.
For example, when a conditional AE builds a model of hand-written digits, the class label $y$ represents the character type (i.e., $0, 1, \dots, 9$) and the latent variable $z$ represents the handwriting characteristics independent of the class label $y$, such as thickness and roundness (see references \cite{Kingma2014a,Sohn2015,Maaloe2015} for more detail).

\subsection{Diagnosis based on Generative Model}\label{sec:generativemodeldiagnosis}
Once the conditional generative model $p_\theta$ is trained, we can assume the class $y$ of a test subject $j$, who has not yet received a diagnosis.
This diagnosis is based on Bayes' rule and the evidence lower bound $\mathcal L(\bm x_j;y)$, which approximates the subject-wise log-likelihood $p_\theta(\bm x_j|y)$.
Specifically, the posterior probability $p(y|\bm x_j)$ of the subject's class $y$ is
\begin{align}
  \begin{split}
  p_\theta(y|\bm x_j)
  &=\dfrac{p_\theta(\bm x_j,y)}{p_\theta(\bm x_j)}\\
  &=\dfrac{p_\theta(\bm x_j,y)}{\sum_{y'\in\{0,1\}} p_\theta(\bm x_j,y')}\\
  &=\dfrac{p(y)p_\theta(\bm x_j|y)}{\sum_{y'\in\{0,1\}} p(y')p_\theta(\bm x_j|y')}\\
  &\approx\dfrac{p(y)\exp\mathcal L(\bm x_j;y)}{\sum_{y'\in\{0,1\}} p(y')\exp\mathcal L(\bm x_j,y')}\\
  &\propto p(y)\exp\mathcal L(\bm x_j;y).
  \end{split}\label{eq:posterior}
\end{align}
Hence, the larger the subject-wise evidence lower bound $\mathcal L(\bm x_j;y)$, the more likely the assumption of the class $y$ of the subject $j$ is correct given the images $\bm x_j$.
In this study, we set the prior probability $p(y)$ of class $y$ to be equal to each other, then
\begin{equation}
\begin{array}{l}
  \hspace*{-3mm}\mathcal L(\bm x_j;y=1)>\mathcal L(\bm x_j;y=0)\\
  \hspace*{5mm}\Leftrightarrow\mbox{the subject $j$ is more likely to have the disorder}\hspace*{-3mm}
\end{array}\label{eq:diagnosis}
\end{equation}
and vice versa.

\begin{figure}[t]\centering
  \includegraphics[page=1,scale=0.4,clip]{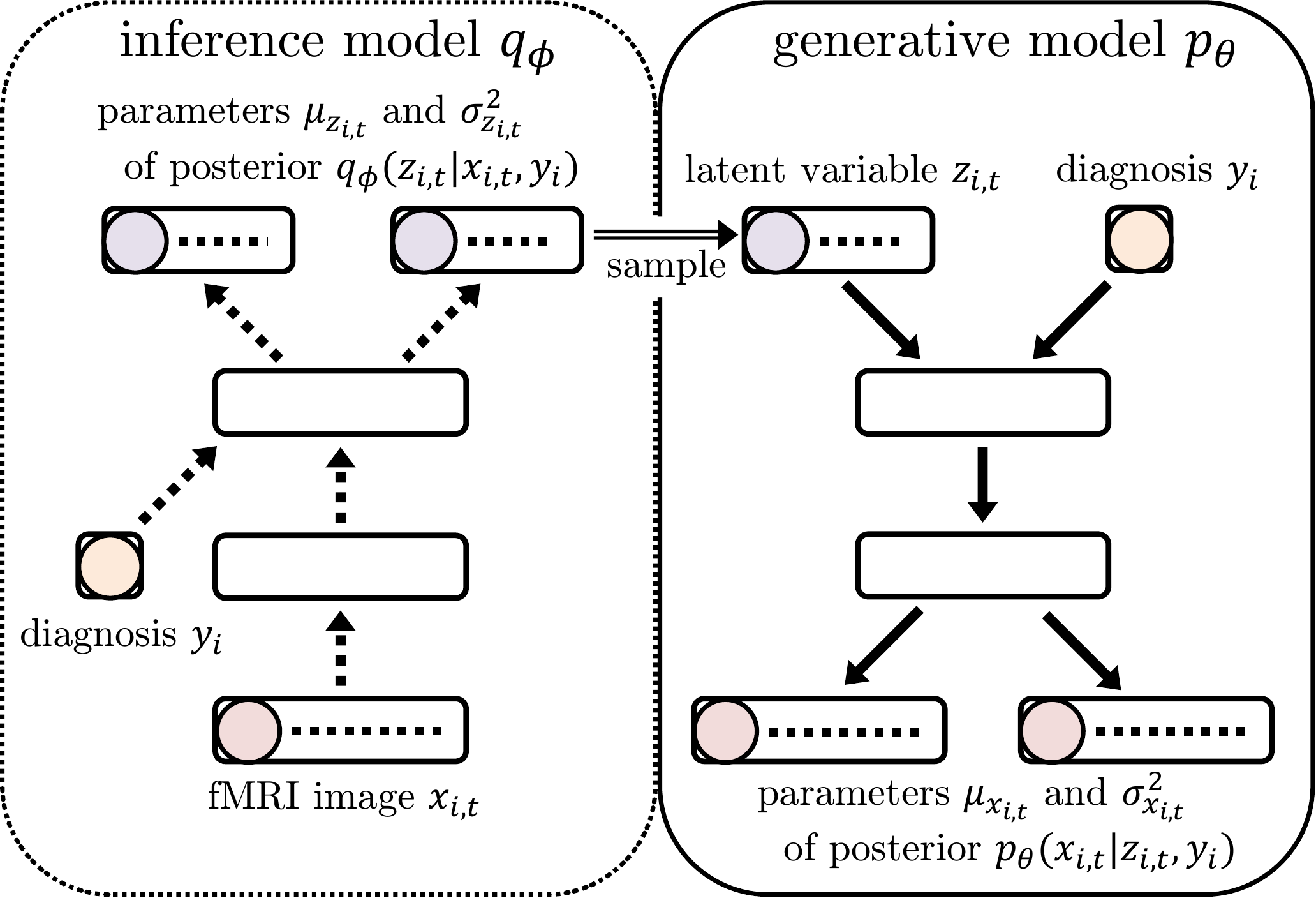}
  \caption{Architectures of deep neural networks representing our proposed frame-wise generative model.}\label{fig:architecture}
\end{figure}

\subsection{Deep Neural Generative Model of FMRI Images}
In this section, we implement the conditional generative model $p_\theta$ described in the previous section using deep neural networks, thus obtaining a deep neural generative model (DGM)~\cite{Kingma2014,Kingma2014a,Sohn2015,Maaloe2015}.
We build and train the frame-wise model $p_\theta(x_{i,t}|y_i)$, and thereby obtain the subject-wise model $p_\theta(\bm x_i|y_i)$ and the model $p_\theta(\bm X|\bm y)$ of the complete dataset.

The inference model $q_\phi(z_{i,t}|x_{i,t},y_i)$ is implemented on a neural network called \emph{encoder}, depicted in the left part of Fig.~\ref{fig:architecture}.
The encoder is given a fMRI feature $x_{i,t}$ and the corresponding class label $y_i$, then infers the posterior distribution $q_\phi(z_{i,t}|x_{i,t},y_i)$ of the latent variable $z_{i,t}$.
Since the posterior distribution $q_\phi(z_{i,t}|x_{i,t},y_i)$ is modeled as a multivariate Gaussian distribution with a diagonal covariance matrix, the encoder outputs a mean vector $\mu_{z_{i,t}}$ and a variance vector $\sigma^2_{z_{i,t}}$.
The conditional generative model $\log p_\theta(x_{i,t}|y_i)$ is implemented on a neural network called \emph{decoder} (or sometimes called \emph{generator}), also depicted in the right part of Fig.~\ref{fig:architecture}.
The decoder is given a class label $y_i$ and a latent variable $z_{i,t}$, then generates the posterior distribution $p_\theta(x_{i,t}|z_{i,t},y_i)$ of an fMRI feature $x_{i,t}$.

More specifically, we constructed the encoder and decoder as follows.
We assumed a fMRI feature $x_{i,t}$ as an $n_x$-dimensional vector, a latent variable $z_{i,t}$ as an $n_z$-dimensional vector, and a class label $y_i$ as a one-hot vector.
The encoder and decoder have $u_h$ hidden layers.
Each hidden layer consists of $n_h$ units followed by layer normalization~\cite{Ba2016} and the ReLU activation function~\cite{Nair2010}.
Each weight parameter was initialized to a sample drawn from a Gaussian distribution $\mathcal N(0,0.02^2)$ and each bias parameter was initialized to 0.
The encoder accepts an fMRI feature $x_{i,t}$ with the dropout~\cite{Srivastava2014} of ratio $p$ at its first hidden layer and a class label $y_i$ at its last hidden layer.
The output layer of the encoder consists of $2\times n_z$ units.
Half of the units are followed by the identity function as an activation function and used as a mean vector $\mu_{z_{i,t}}$, and the other half of the units are followed by the exponential function as an activation function and used as a variance vector $\sigma^2_{z_{i,t}}$.
The two vectors represent the parameters of the variational posterior $q_\phi(z|x_{i,t})=\mathcal N(\mu_{z_{i,t,}},\mathrm{diag}(\sigma^2_{z_{i,t}}))$, which is a $n_z$-dimensional multivariate Gaussian distribution with a diagonal covariance matrix.
We assumed the prior distribution $p(z)$ to be a multivariate standard Gaussian distribution.
Then, we can calculate the Kullback-Leibler divergence $D_{KL}(q_\phi(z_{i,t}|x_{i,t},y)||p(z))$ in Eq.~\eqref{eq:scanlikelihood} and sample a latent variable $z_{i,t}$.
The decoder accepts a sample $z_{i,t}$ from the variational posterior $q_\phi(z_{i,t}|x_{i,t},y_i)$ and the class label $y_i$ at its first hidden layer.
The output layer of the decoder consists of $2\times n_x$ units.
As is the case with the encoder, half of the units are followed by the identity function as an activation function and used as a mean vector $\mu_{x_{i,t}}$.
The other half of the units are followed by the exponential function as an activation function and used as a variance vector $\sigma^2_{x_{i,t}}$.
The two vectors represent the parameters of the posterior distribution $p_\theta(x_{i,t}|z_{i,t},y_i)=\mathcal N(\mu_{x_{i,t}},\mathrm{diag}(\sigma^2_{x_{i,t}}))$, which is a $n_x$-dimensional multivariate Gaussian distribution with a diagonal covariance matrix.
Then, we can calculate the log-likelihood $\log p_\theta(x_{i,t}|z_{i,t},y_i)$ in Eq.~\eqref{eq:scanlikelihood}.

The encoder and decoder were jointly trained using the Adam optimization algorithm~\cite{Kingma2014b} with parameters $\alpha=10^{-4}$, $\beta_1=0.9$, and $\beta_2=0.999$.
We selected hyper-parameters from $p\in\{0.0,0.5\}$, $n_h\in\{100,200,400\}$, and $n_z\in\{5,10,20,50,100\}$ for $n_h>n_z$.
Note that, while deeper and deeper convolutional and recurrent neural networks are attracting increasing attention (e.g., \cite{He2015a,Bahdanau2014}), recent state-of-the-art feedforward fully-connected neural networks have one or two hidden layers~\cite{Kingma2014,Kingma2014a,Sohn2015,Maaloe2015}, and a deeper network architecture is not always helpful~\cite{Noekland2016}.
Hence, we set the number $u_h$ of hidden layers to two.
Since $z_{i,t}$ is a continuous random variable, the expectation $\mathbb E_{q_\phi(z_{i,t}|x_{i,t},y_i)}\left[\log p_\theta(x_{i,t}|z_{i,t},y_i)\right]$ is calculated by Monte Carlo sampling from the variational posterior $q_\phi(z|x)$.
Following the original implementation~\cite{Kingma2014,Kingma2014a,Sohn2015}, the latent variable $z$ was sampled once per iteration during the parameter adjustment.
In the evaluation, the maximum a posteriori (MAP) estimate of the latent variable $z$ (i.e., the mean vector $\mu_{z}$) was used in place of Monte Carlo sampling from the variational posterior $q_\phi(z|x)$.
As is the case with the original implementation, we confirmed that this simplification does not have an obvious negative effect on the accuracy.
We adjusted the imbalance in the classes via oversampling; hence, we assumed the prior probabilities $p(y)$ of classes $y$ as $p(y=0)=p(y=1)=0.5$.

Basically, we stopped the learning procedure if the evidence lower bound $\mathcal L(\bm X;y)$ shown in Eq.~\eqref{eq:wholelikelihood} converged.
Since our purpose is diagnosing subjects, we additionally employed the following early-stopping criterion in a supervised learning manner~\cite{Prechelt2012}.
During the training procedure, we evaluated the diagnostic accuracy on the training subjects using Eq.~\eqref{eq:posterior} every 100 iterations and selected the results when the training accuracy reached the best.

We implemented our proposed DGM and comparative approaches using Python v3.6.3 with libraries, TensorFlow v1.8.0~\cite{tensorflow} and scikit-learn v0.19.1~\cite{Pedregosa2011}.

\section{Comparative Approaches}\label{sec:comparative}

\subsection{Model of Functional Connectivity}\label{sec:model_connectivity}
For comparison, we evaluated two diagnostic methods based on functional connectivity (FC).
FC is connection between regions that function together, including via indirect connections without underlying structural connectivity.
The FCs of a subject $i$ have been typically measured by Pearson correlation coefficients (PCCs) between the region-wise fMRI signals $x_i$.
Many studies have reported that changes in FC are associated with psychiatric disorders (see \cite{Castellanos2013,Watanabe2017} for a review).
The PCC between $n_x$ regions provides a $\frac{1}{2}n_x(n_x-1)$-dimensional FC vector $c_i$ per subject.
Since the FC vector $c_i$ is of high-dimension, feature-extractors are required.
The selected FCs can be considered as the FCs related to the disorder.

Following~\cite{Shen2010,Tang2012}, we employed the Kendall $\tau$ rank correlation coefficient to choose FCs.
We selected $m$ FCs of the largest correlations with the class label $y_i$.
Subsequently, we employed locally linear embedding (LLE) with a parameter $k$ to project the selected FCs to a $d$-dimensional space.
Finally, the c-means clustering algorithm classified the $k$-dimensional features.
The c-means clustering employed the arccosine of the cosine similarity as the dissimilarity.
The original study~\cite{Shen2010} confirmed that this procedure outperformed direct classifications of the FC vector $c_i$ by the SVM and MLP.
Henceforth, we denote this procedure as PCC+Kendall+LLE+c-means.
We selected the number $m$ of the selected feature from $m\in\{50,100,200,400,600\}$, the parameter $k$ of the LLE from $k\in\{5,8,10,12,15\}$, and the dimension number $d$ of the final features from $d\in\{1,2,5,10,20,50\}$, where these ranges followed the original study~\cite{Shen2010}.

Additionally, following~\cite{Yahata2016,Takagi2017}, we employed an $L_1$-regularized sparse canonical correlation analysis (SCCA) to reduce the risk of extracting disorder-unrelated signals from the FC vector $c_i$.
Canonical correlation analysis is a linear generative model, in which the two visible variables have private latent variables and share a latent variable called a canonical variable.
To permit a fair comparison, we only used the FC vector $c_i$ and class label $y_i$ as the visible variables of the SCCA, but we could add other attributes such as age and gender~\cite{Yahata2016,Takagi2017}.
We selected $d$ elements in the FC vector $c_i$ that had connections to the canonical variable with large weights.
Finally, we employed sparse logistic regression (SLR) to classify the remaining FCs.
The sparsity was determined by automatic relevance determination (ARD)~\cite{Neal1996}.
Henceforth, we denote this procedure as PCC+SCCA+SLR.
We selected the number $d$ of the selected feature from $d\in\{50,100,200,400,600\}$.

\subsection{Model of Individual Frames}\label{sec:model_frame}
We also evaluated several models that consider individual frames, as in our proposed DGM, for comparison.

A multilayer perceptron (MLP) is a feedforward neural network that acts as a classifier.
The MLP accepted a single image $x_{i,t}$ simultaneously.
It had $u_h$ hidden layers, each of which consisted of $n_h$ units followed by layer normalization~\cite{Ba2016} and the ReLU activation function~\cite{Nair2010} as is the case in our proposed DGM.
The MLP also had an output unit followed by the logistic function, representing the posterior probability $q_\phi(y=1|x_{i,t})$.
The objective function to be minimized was cross-entropy $\mathcal L_{c.e.}=- \sum_{y} \mathbb I(y=y_i)\log q_\phi(y|x_{i,t})$, where $\mathbb I(cond.)$ is the indicator function that returns 1 if $cond.$ is true and 0 otherwise.
The other conditions were the same as those for our proposed DGM.
Once the MLP was trained, it sequentially accepted a set $\bm x_j=\{x_{j,t}\}$ of fMRI features obtained from a subject $j$ and diagnosed the subject using the ensemble of the diagnoses for the $T_j$ images, also consistent with our proposed DGM.
Then, $\sum_{t=1}^{T_j} \log q_\phi(y=1|x_{j,t})>\sum_{t=1}^{T_j} \log q_\phi(y=0|x_{j,t})$ was considered to suggest that subject $j$ had the disorder $y=1$ and vice versa.

For unsupervised feature-extractor, we employed an autoencoder (AE).
We selected hyper-parameters of AE from $p\in\{0.0,0.5\}$, $u_h=2$, $n_h\in\{50,100,200,400\}$, and $n_z\in\{5,10,20,50,100\}$ for $n_h>n_z$, consistent with the proposed DGM.
We used mean-squared-errors for evaluating reconstruction by AE.
The other conditions for AE were the same as those in our proposed DGM and the MLP.
For the combination of the AE and MLP, we used the same number $n_h$ of hidden units for both the AE and MLP in order to suppress its relatively high dimensional hyperparameter-space.

The generative model of fMRI features described in Section~\ref{sec:generativemodel} can be implemented using other generative models.
For comparison, we evaluated a GMM with full covariance matrices~\cite{Murphy2012}.
This GMM can be considered a single-layer version of the proposed DGM but it has a discrete latent variable $z$ and is trained using the Expectation-Maximization (EM) algorithm.
We trained two GMMs $p_\theta(x_{i,t}|y=1)$ and $p_\theta(x_{i,t}|y=0)$: one for patients with disorder ($y=1$) and the other for normal control subjects ($y=0$).
Then, we diagnosed the subjects as described in Section~\ref{sec:generativemodeldiagnosis}, also consistent with our proposed DGM.
We selected the number $n$ of mixture components of the GMM from $n\in\{2,5,10,20,50,100\}$.

\subsection{Model of Dynamic Functional Connectivity}\label{sec:model_temporal}
Recent studies have revealed that the functional connectivity varies over time and is called dynamic FC~\cite{Liu2013,Li2015f}.
We evaluated a hidden Markov model (HMM)~\cite{Murphy2012} and long short-term memory (LSTM)~\cite{Hochreiter1997}.
As distinct from the aforementioned models, HMM and LSTM consider the temporal dynamics of the latent variable and potentially capture the dynamic FCs~\cite{Castellanos2013,Suk2015,Suk2016,Sourty2016,Nielsen2016a,Vidaurre2017}.

Unlike the vanilla GMM, the HMM infers the latent variable $z_{i,t}$ at time $t$ on the basis not only of the observed signal $x_{i,t}$ but also of the estimate of the last latent variable $z_{i,t-1}$.
Following \cite{Suk2016}, we examined the HMM in a procedure based on likelihood like our proposed DGM and the GMM.
We selected the number $n$ of mixture components of the HMM from $n\in\{2,5,10,20,50,100\}$.

We also examined a procedure based on the dynamic FC extracted by an HMM~\cite{Vidaurre2017,Sourty2016}.
We trained an HMM with $n$ states using all subjects for training.
Then, the HMM built multiple templates of FCs as there base distributions $p(x|z)$ corresponding to the hidden states $z$, and a subject belongs to one of FCs (hidden states) at each frame dynamically.
Hence, this method can be considered to capture the dynamic FCs.
Using the HMM, we inferred the posterior probability $p(z_{i,t}=k|\bm x_i)$ that a subject $i$ belonged to a hidden state $k\in\{1,\dots,n\}$ at each frame $t$, and averaged the probability over the time, obtaining the fractional occupancy (FO) $\mathbb E_t[p(z_{i,t}|\bm x_i)]$ of the states for the subject $i$.
We classified the FO as a feature of a subject using SVM.
Henceforth, we denote this procedure as HMM(FO)+SVM.
We selected the number $n$ of mixture components of the HMM from $n\in\{2,5,10,20,50,100\}$ and the hyper-parameter $C$ from $C\in\{\dots,0.1,0.2,0.5,1,2,5,10,\dots\}$ for the SVM to trade-off between the classification accuracy and margin maximization.

The LSTM is a recurrently-connected neural network with specially designed units.
It has already been employed in previous work~\cite{Dvornek2017}.
The LSTM accepted images $x_{i,t}$ sequentially and outputted a diagnosis $y$ at the last time step $T_i$.
We set the number of hidden layers to one, and we selected the number $n_h$ of hidden units from $n_h\in\{50,100,200,400\}$ and the dropout ratio from $p\in\{0.0,0.5\}$.

\section{Experiments and Results}\label{sec:results}

\subsection{Data Acquisition and Preprocessing}\label{sec:preprocessing}
Actually, our proposed DGM has a general-purpose structure, which accepts any types of fMRI time-series.
Each fMRI feature $x_{i,t}$ can be a 3D image, a 2D image, a k-space image, a vector of voxels, a feature vector of regions-of-interest (ROIs), or a state of dynamic functional connectivity.
In this study, we evaluate our proposed DGM on vectors of ROI-wise features.
This is because many comparative studies have been conducted on the ROI-wise features~\cite{Shen2010,Tang2012,Suk2016,Suk2017,Yahata2016,Zhang2005a,Plis2014,Vergun2016,Castro2016,Watanabe2017,Dvornek2017}.
In this situation, we can identify the names of ROIs contributing to the disorder, which is one of the main concerns of studies on fMRI data (see also Section~\ref{sec:contribution}).
Application of our proposed method to a time-series of unpreprocessed fMRI 3D images is a potential future work but out of scope of this study.

In this study, we used a dataset of rs-fMRI images obtained from patients with schizophrenia or bipolar disorder.
These data were obtained from the OpenfMRI database.
Its accession number is ds000030 (\url{https://openfmri.org/dataset/ds000030/}).
We used all available subjects in the dataset: 50 patients with schizophrenia, 49 patients with bipolar disorder, and 122 normal control subjects.
The environmental settings were repetition time (TR) $=$ 3000 ms, acquisition matrix size $=$ $64\times64\times34$, 152 frames, and voxel thickness $=$ 3.0 mm.
We employed the preprocessed version, in which time-slice adjustment, rigid body rotation to correct for displacement, and spatial normalization to the MNI space were already performed following the fMRIprep pipeline~\cite{Esteban2019} (see also \url{https://github.com/poldracklab/fmriprep}).
As data scrubbing, we discarded frames with framewise displacements (FD) of more than 1.5 mm or angular rotations of more than 1.5 degrees in any direction as well, as the subsequent frames.
We also discarded the data of subjects who had fewer than 100 consecutive frames remaining.
Note that, following the previous study~\cite{Suk2016}, we set the threshold for discarding frames slightly larger than some other studies~\cite{Yahata2016}.
This is because we examine the models (HMM and LSTM) that require a time-series (consecutive frames) of fMRI images.
With a lower threshold, we obtain only a limited number of remaining time-series and it is difficult to train these models.

We parcellated each fMRI image into 116 ROIs using an automated anatomical labeling (AAL) template \cite{Tzourio-Mazoyer2002}.
The AAL template is one of the most commonly used templates~\cite{Shen2010,Tang2012,Suk2016,Yahata2016,Castro2016,Watanabe2017,Suk2017}.
The data of subjects whose fMRI images did not match the template even after the spatial normalization were also discarded.
We averaged voxel intensities in each ROI to obtain 116 dimensional vectors of ROI-wise intensities.
Finally, we bandpass-filtered each time series of ROI-wise intensity to the frequency range between 0.01 Hz and 0.1 Hz and normalized it to zero mean and unit variance.

As a result, we obtained 117 normal subjects, 48 patients with schizophrenia, and 46 patients with bipolar disorder.
Hence, the parameters of the datasets were $N=165$ for the schizophrenia dataset, $N=163$ for the bipolar disorder.
While some studies have proposed methods to correct displacements without discarding frames, fMRIprep is not the case and discarding frames is still a popular step in preprocessing of fMRI data~\cite{Yahata2016,Suk2016}.
After scrubbing, only 4 normal control subjects, 2 patients with schizophrenia, and no patients with bipolar disorder had fewer frames.
Since we discarded the data of subjects who had fewer than 100 consecutive frames remaining, the imbalance in the number of frames was less than 1.5.
Moreover, throughout this study, we adjusted the imbalance of frames per subject and that of subjects per class for training models (see Sections~\ref{sec:methods} and \ref{sec:comparative}).

\begin{table*}[t]\centering
  \caption{Selected Hyper-Parameters and Diagnostic Accuracies for Schizophrenia. (Preprocessed)}\label{tab:results_schizophrenia}
  \setlength{\tabcolsep}{3pt}
  \begin{tabular}{lccccc}
    \toprule
      \multirow{2}{*}{\textbf{Model}} & \multicolumn{2}{c}{\textbf{Selected Hyper-Parameters}} & \multicolumn{3}{c}{\textbf{Balanced Measures}}\\
      \cmidrule(lr){2-3}\cmidrule(lr){4-6} &
      \textbf{Feature-Extractor} & \textbf{Classifier} & \textbf{BACC} & \textbf{MCC} & \textbf{F1}\\
    \midrule
      chance level  & --- & ---
      & 0.500 & 0.000   & 0.000  \\
    \midrule
      PCC+Kendall+LLE+c-means~\cite{Shen2010}     & $m\!=\!50$, $k\!=\!12$, $d\!=\!5$& (no parameter)
      & 0.661 & 0.299 & \underline{0.535}\\
      PCC+SCCA+SLR~\cite{Yahata2016}     & $d=600$ & (no parameter)
      & 0.664 & 0.335 & 0.512\\
    \midrule
      GMM     & ---   & $n = 20$
      & 0.605 & 0.222 & 0.425\\
      MLP           & --- & $n_h=400$, $p=1.0$
      & 0.590 & 0.257 & 0.256\\
      AE+MLP        & $n_h\!=\!400$, $n_z\!=\!50$, $p\!=\!0.5$ & $n_h=400$, $p=1.0$
      & \underline{0.679} & \underline{0.423} & 0.504 \\
    \midrule
      HMM~\cite{Suk2016}           & --- & $n = 10$
      & 0.580 & 0.150 & 0.430\\
      HMM(FO)+SVM~\cite{Vidaurre2017}           & $n=5$ & $C = 20$
      & 0.632 & 0.279 & 0.457\\
      LSTM~\cite{Dvornek2017}           & --- & $n_h=100$, $p=1.0$
      & 0.664 & 0.349 & 0.513\\
    \midrule
      DGM (proposed) & --- & $n_h=100$, $n_z=5$, $p=1.0$
      & \textbf{0.713} & \textbf{0.438} & \textbf{0.581}\\
    \bottomrule
  \end{tabular}
  \vspace*{1.0cm}
  \caption{Selected Hyper-Parameters and Diagnostic Accuracies for Bipolar Disorder. (Preprocessed)}\label{tab:results_bipolar}
  \setlength{\tabcolsep}{3pt}
  \begin{tabular}{lcccccccccc}
    \toprule
      \multirow{2}{*}{\textbf{Model}} & \multicolumn{2}{c}{\textbf{Selected Hyper-Parameters}} & \multicolumn{3}{c}{\textbf{Balanced Measures}}\\
      \cmidrule(lr){2-3}\cmidrule(lr){4-6} &
      \textbf{Feature-Extractor} & \textbf{Classifier} & \textbf{BACC} & \textbf{MCC} & \textbf{F1}\\
    \midrule
      chance level  & --- & ---
      & 0.500 & 0.000 & 0.000 \\
    \midrule
      PCC+Kendall+LLE+c-means~\cite{Shen2010}     & $m\!=\!200$, $k\!=\!12$, $d\!=\!2$ & (no parameter)
      & \underline{0.622} & \underline{0.226} & \underline{0.490} &\\
      PCC+SCCA+SLR~\cite{Yahata2016}     & $d=600$ & (no parameter)
      & 0.599 & 0.206 & 0.397 \\
    \midrule
      GMM     & ---   & $n = 2$
      & 0.523 & 0.085 & 0.083 \\
      MLP           & --- & $n_h=50$, $p=0.5$
      & 0.531 & 0.098 & 0.101 \\
      AE+MLP         & $n_h\!=\!400$, $n_z\!=\!100$, $p\!=\!0.5$ & $n_h=400$, $p=1.0$
      & 0.571 & 0.174 & 0.243 \\
    \midrule
      HMM~\cite{Suk2016}           & --- & $n=10$
      & 0.548 & 0.144 & 0.197 \\
      HMM(FO)+SVM~\cite{Vidaurre2017}           & $n=2$ & $C = 100$
      & 0.569 & 0.141 & 0.404 \\
      LSTM~\cite{Dvornek2017}           & --- & $n_h=200$, $p=0.5$
      & 0.585 & 0.190 & 0.384 \\
    \midrule
      DGM (proposed) & --- & $n_h\!=400$, $n_z\!=100$, $p\!=0.5$
      & \textbf{0.640} & \textbf{0.278} & \textbf{0.491} \\
    \bottomrule
  \end{tabular}
\end{table*}

\subsection{Results of Diagnosis}\label{sec:accuracy}
Let $\mathrm{TP}$, $\mathrm{TN}$, $\mathrm{FP}$, and $\mathrm{FN}$ denote true positive, true negative, false positive, and false negative, respectively.
Then, we introduce several measures, namely accuracy $(\mathrm{ACC})$, sensitivity $(\mathrm{SEN})$, specificity $(\mathrm{SPEC})$, positive predictive value $(\mathrm{PPV})$, and negative predictive value $(\mathrm{NPV})$, defined as
\begin{align*}
  \mathrm{ACC}&=\mathrm{(TP+TN)/(TP+TN+FP+FN)},\\
  \mathrm{SEN}&=\mathrm{TP / (TP + FN)},\\
  \mathrm{SPEC}&=\mathrm{TN / (TN + FP)},\\
  \mathrm{PPV}&=\mathrm{TP / (TP + FP)},\\
  \mathrm{NPV}&=\mathrm{TN / (TN + FN)}.
\end{align*}
While these measures show the tendencies of models, they are inappropriate for performance evaluations.
Recall that the datasets contain many more controls than patients; that is, the datasets are highly imbalanced.
A model can readily obtain a large score on these measures with a biased prediction, e.g., a model estimating all subjects as controls has an accuracy of 71 \% and a specificity of 100 \%.
Hence, we primarily used the following balanced measures:
balanced accuracy $(\mathrm{BACC})$, Matthews correlation coefficient $(\mathrm{MCC})$, and F$_1$ score $(\mathrm{F1})$ defined as
\begin{align*}
  \mathrm{BACC}&=\mathrm{(SEN+SPEC)/2},\\
  \mathrm{MCC} &=\frac{\mathrm{(TP\times TN-FP\times FN)}}{\mathrm{(TP+FP)(TP+FN)(TN+FP)(TN+FN)}},\\
  \mathrm{F1}  &=\mathrm{2/(SEN^{-1}+PPV^{-1})}.
\end{align*}
These measures are robust to the imbalance in datasets and appropriate for performance evaluations.
We performed 10 trials of 10-fold cross-validations and selected hyper-parameters with which the models achieved the best balanced accuracy ($\mathrm{BACC}$), as summarized in Table~\ref{tab:results_schizophrenia} for schizophrenia and in Table~\ref{tab:results_bipolar} for bipolar disorder.
The proposed DGM achieved the best results for all the balanced measures by obvious margins, except for PCC+Kendall+LLE+c-means~\cite{Shen2010} in the F$_1$ score for the bipolar disorder dataset.
The best results are emphasized in bold, and the second-best results are emphasized by underlines.
We also summarize the results of non-balanced measures in Table~\ref{tab:appendresults} in the Appendix, purely for reference.

\subsection{Reconstruction of Signals and Contributing Regions}\label{sec:contribution}
In the previous section, we diagnosed subjects successfully following Eq.~\eqref{eq:diagnosis}, i.e., using the difference in the subject-wise evidence lower bound $L(\bm x_j;y)$ between the given class labels $y$.
In this section, we visualize the regions contributing to the diagnoses.
The frame-wise evidence lower bound $\mathcal L(x_{i,t};y)$ is equal to the log-likelihood $\mathbb E_{q_\phi}[\log p_\theta(x_{i,t}|z_{i,t},y)]$ of an ROI-wise feature $x_{i,t}$ minus the Kullback-Leibler divergence $D_{KL}(q_\phi(z_{i,t}|x_{i,t},y)||p(z_{i,t}))$.
From the perspective of a neural network, the former is the negative reconstruction error of an autoencoder and the latter is a regularization term~\cite{Kingma2014,Kingma2014a,Sohn2015,Maaloe2015}.
Here, we explicitly denote an ROI-wise feature $x_{i,t}$ as the set of the region-wise signals $x_{i,t,k}$ for $k\in\{1,2,\dots,116\}$ and introduce a region-wise reconstruction error given a class label $y$:
\begin{align}
\begin{split}
  \mathcal E(i,t,k;y)\hspace*{-15mm}\\
  &=-\mathbb E_{q_\phi(z_{i,t}|x_{i,t},y)}\left[\log p_\theta(x_{i,t,k}|z_{i,t},y)\right]\\
  &=-\mathbb E_{q_\phi(z_{i,t}|x_{i,t},y)}\left[\log \frac{1}{\sqrt{2\pi\sigma_{x_{i,t}}^2}}\exp\left(-\frac{|x_{i,t}-\mu_{x_{i,t}}|^2}{2\sigma_{x_{i,t}}^2}\right)\right]\\
  &=\mathbb E_{q_\phi(z_{i,t}|x_{i,t},y)}\left[\log\sigma_{x_{i,t}}+\frac{|x_{i,t}-\mu_{x_{i,t}}|^2}{2\sigma_{x_{i,t}}^2}\right]+\frac{1}{2}\log2\pi.
\end{split}
\end{align}
When the reconstruction error of a region $k$ becomes much larger given the incorrect class label, the proposed DGM disentangles the signals of the region $k$ obtained from controls and patients and the region $k$ contributes largely to the correct diagnosis.

\begin{figure}[t]
  \centering
  \includegraphics{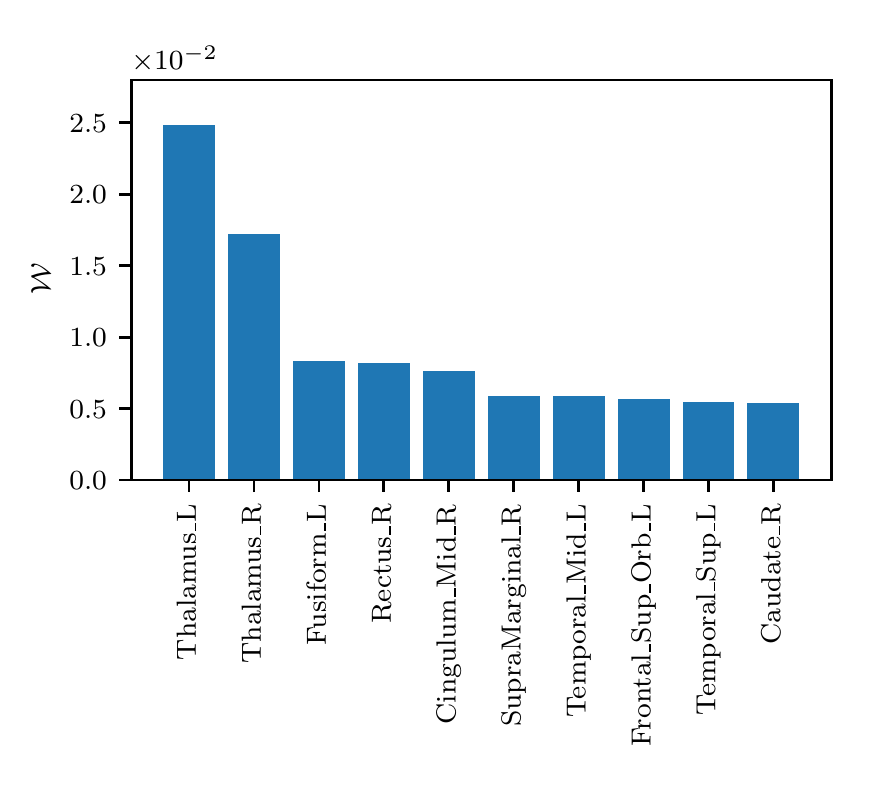}
  \vspace*{-5mm}
  \caption{Top 10 contributing regions and their contribution weights for schizophrenia, defined in Eq.~\eqref{eq:contribution}.}
  \label{fig:contribution_schizophrenia}
  \vspace*{-5mm}
\end{figure}

\begin{figure}[t]
  \centering
  \includegraphics{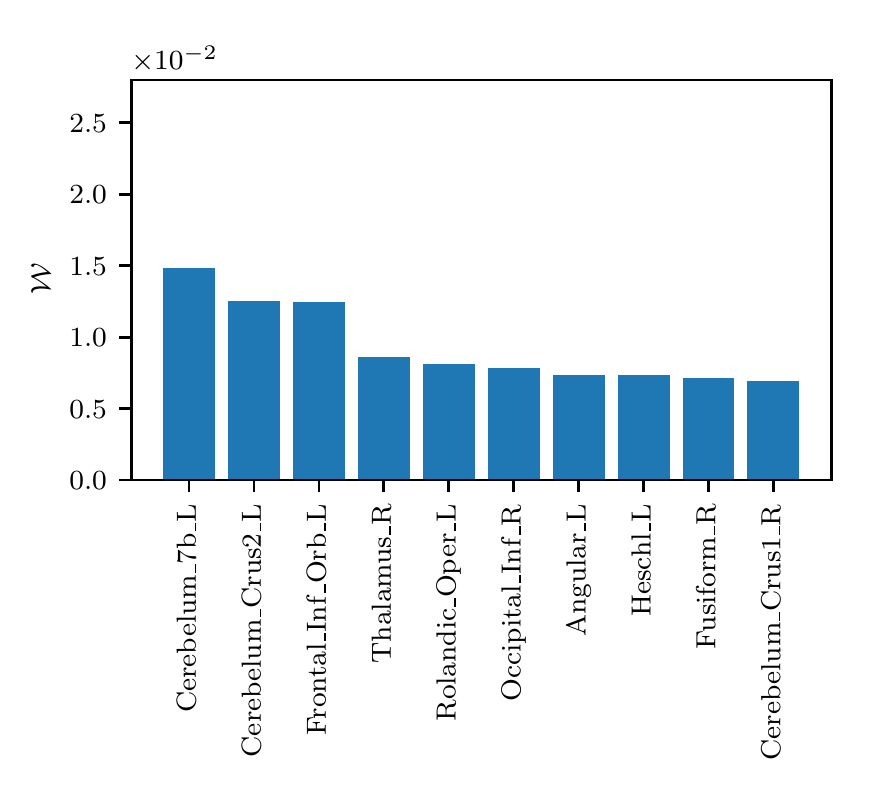}
  \vspace*{-5mm}
  \caption{Top 10 contributing regions and their contribution weights for bipolar disorder, defined in Eq.~\eqref{eq:contribution}.}
  \label{fig:contribution_bipolar}
  \vspace*{-5mm}
\end{figure}

We define the contribution weight $\mathcal W(i,k,t)$ of a region $k$ at a frame $t$ obtained from subject $i$ as
\begin{equation}
    \mathcal W(i,t,k)=\mathcal E(i,t,k;y\!=\!1\!-\!y_i)-\mathcal E(i,t,k;y\!=\!y_i),\label{eq:framecontribution}
\end{equation}
By averaging over frames and subjects, we define the contribution weight $\mathcal W(k)$ of a region $k$ as
\begin{equation}
    \mathcal W(k)=\mathbb E_{i,t} [\mathcal W(i,t,k)],\label{eq:contribution}
\end{equation}
where the imbalances in subjects per class and frames per subject are adjusted.
Recall that $y_i$ denotes the correct class label of the subject $i$ and $1-y_i$ denotes the incorrect class label.
We summarize the regions with the top 10 largest contribution weights in Figs.~\ref{fig:contribution_schizophrenia} and \ref{fig:contribution_bipolar}.

\begin{figure*}[t]
  \centering
  \includegraphics{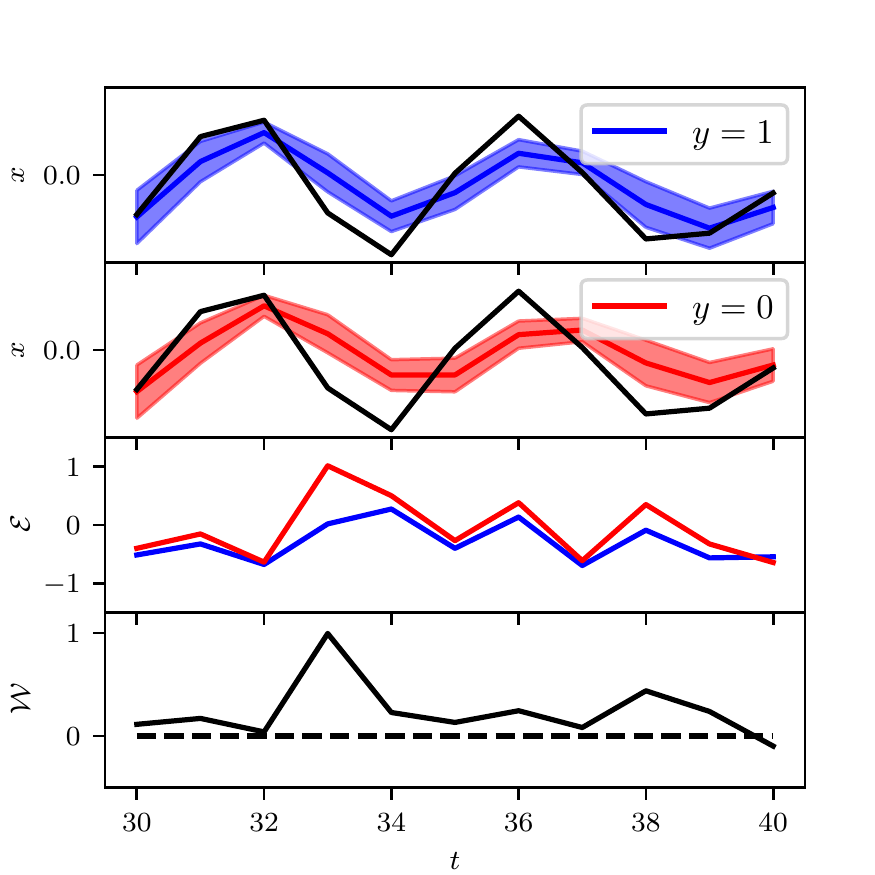}
  \includegraphics{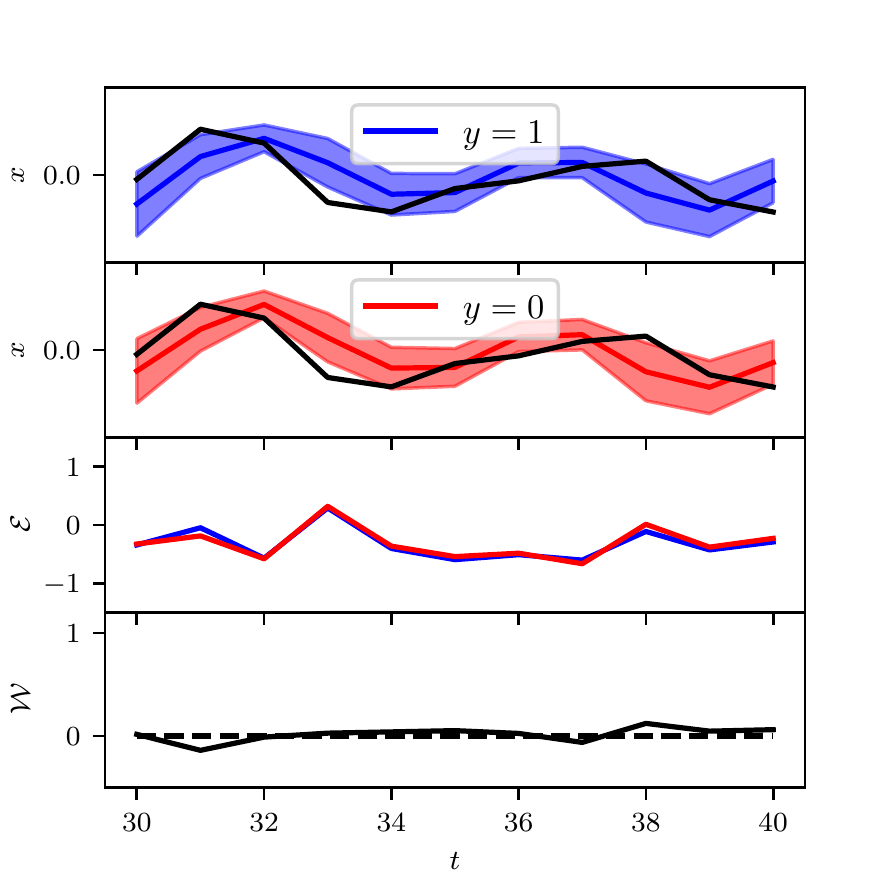}
  \vspace*{-1mm}
  \caption{
    The time-series of the signals $x_{i,t,k}$ and reconstruction errors $\mathcal E(i,t,k;y)$ of the left thalamus (left panel) and the left anterior cingulum (right panel) of a subject with schizophrenia ($y_i=1$).
    The black lines denote the obtained fMRI signals.
    The colored lines with shaded areas denote the mean and standard deviation of the posterior distribution $p(x_{i,t,k}|z_{i,t},y)$ of the signal, where the blue and red colors correspond to the correct label $y=y_i=1$ and the incorrect label $y=1-y_i=0$, respectively.
    The colored lines in the second bottom panels denote the region-wise reconstruction error $\mathcal E(i,t,k;y)$ after the constant bias $\frac{1}{2}\log2\pi$ was subtracted.
    The black lines in the bottom panels denote the region-wise contribution weight $\mathcal W(i,t,k)$.
  }
  \label{fig:timeseries}
  \vspace*{-5mm}
\end{figure*}

Additionally, we visualize the time-series of the signal $x_{i,t,k}$ obtained from a subject with schizophrenia in Fig.~\ref{fig:timeseries} by the black lines.
In the two top panels, we denote the posterior probability of the signals given the correct label $y=1$ and incorrect label $y=0$ by the blue and red lines.
We also denote the region-wise reconstruction errors $\mathcal E(i,t,k;y)$ and contribution weights $\mathcal W(i,t,k)$ in the two bottom panels.

For these visualizations, we used the log-likelihood $\log p_\theta(x_{i,t}|z=\mu_z,y_i)$ given the MAP estimate $\mu_z$ of the latent variable $z$ instead of the exact log-likelihood $\mathbb E_{q_\phi(z_{i,t}|x_{i,t},y_i)}\left[\log p_\theta(x_{i,t}|z_{i,t},y_i)\right]$.
This is because we also used the MAP estimate $\mu_z$ for diagnosis in the previous sections and we found no obvious difference from the approximation of the integral using Monte Carlo sampling.

\section{Discussion}\label{sec:discussion}
\subsection{Diagnostic Accuracy}
We summarize the selected hyper-parameters and results in Tables~\ref{tab:results_schizophrenia} and \ref{tab:results_bipolar}.
While the performance of each method depends on the datasets, our proposed DGM achieved the best results for both datasets and for all measures.
While the PCC+Kendall+LLE+c-means procedure provides a competitive $F_1$ score to our proposed DGM for the bipolar disorder dataset, its result for the schizophrenia dataset is clearly worse than ours.
The AE+MLP also provides the second best result for the schizophrenia dataset but not for the bipolar disorder dataset.
Hence, our proposed DGM could be unifiedly applicable to many types of psychiatric disorders.

Moreover, the significance is not limited to the classification accuracy.
Both the second best methods employed the dimensional-reduction methods (LLE and AE), which result in compressed features in a lower-dimensional space and have a difficulty in interpreting the compressed features.
Our proposed DGM did not perform dimension-reductions, and thus identifies input elements contributing the diagnosis (see Section~\ref{sec:contribution} for more details).
This is a remarkable advantage compared to the second best methods.

By comparing between MLP and AE+MLP, the unsupervised feature-extraction worked well.
The HMM and LSTM consider temporal dynamics and improved the diagnostic accuracies on average compared to their non-temporal counterparts, the GMM and MLP.
While the proposed DGM neither employ a feature-extractor nor model the temporal dynamics explicitly, it outperformed these models.

The absolute balanced accuracy was not significant when compared with the results reported in previous studies.
This study and previous studies have evaluated models using cross-validation~\cite{Shen2010,Tang2012,Suk2016,Suk2017,Dvornek2017}.
Cross-validation reports the best validation result, which tends to be unreasonably good by chance~\cite{Cawley2010,Yahata2016}.
Especially, the risk becomes higher when the number of hyperparameter candidates is large and the number of samples is small.
We evaluated the hyperparameters within the same range as the original studies sparsely instead of evaluating all candidates one-by-one.
We employed one of the biggest fMRI datasets and reported the average scores of 10 trials.
These conditions made our results more reliable and surpass an unreasonably good result.
While a nested cross-validation provides a more reliable numerical evaluation, it requires vast computational time and is still uncommon among studies on fMRI~\cite{Cawley2010,Yahata2016}.
The current cross-validation is enough for model comparison~\cite{Wainer2018}.

\subsection{Functional Connectivity in Proposed DGM}\label{sec:dgmfc}
The functional connectivities (FCs) of a subject have been measured by Pearson correlation coefficients (PCCs) of the fMRI signals between ROIs~\cite{Castellanos2013,Watanabe2017,Yahata2016,Takagi2017,Shen2010,Tang2012}.
The GMM is literally a mixture of Gaussian distributions. The GMM built a model of FCs between ROIs as full covariance matrices, wherein the covariance was equivalent to the PCC since each fMRI signal was normalized to zero mean and unit variance.
More specifically, the pair of GMMs $p_\theta(x_{i,t}|y=1)$ and $p_\theta(x_{i,t}|y=0)$ built $n$ prototypical FCs of the patients $y=1$ and normal control subjects $y=0$, where $n$ is the number of mixture components.
The diagnosis of a subject is based on the log-likelihood of the subject's fMRI signals, that is, the similarity of the subject's FCs to the prototypes.
Similarly, our proposed DGM built a model $p_\theta(x|y)$ of FCs.
Although given the class label $y$ and the frame-wise variability $z$ the posterior $p_\theta(x|y,z)$ has a diagonal covariance matrix, the posterior $p_\theta(x|y)=\int_z p_\theta(x|y,z)$ only given the class label $y$ could have a full covariance matrix and higher-order correlations thanks to the non-zero mean and the integral and the nonlinearity of the decoder.
The diagnosis is also based on the similarity of the subject's FCs to the prototypes $p_\theta(x|y)$ of the conditions $y$.
Hence, the GMM and our proposed DGM built models of FCs and use them for diagnosis in a different way from the PCCs.

The PCCs, GMM, and proposed DGM do not take into account the temporal order of the fMRI signals.
Recent studies have revealed that the functional connectivity varies over time and is called dynamic FC~\cite{Liu2013,Li2015f}.
Many recent studies have addressed the dynamic FC by employing HMMs and LSTMs~\cite{Suk2015,Suk2016,Sourty2016,Nielsen2016a,Vidaurre2017,Dvornek2017}, where the order of the fMRI signals matters.
According to Tables~\ref{tab:results_schizophrenia} and \ref{tab:results_bipolar}, temporal dynamics improves the diagnostic accuracies on average.
However, we consider that our proposed DGM is appropriate as a first step of application of DGMs to fMRI data analysis because it rivaled or surpassed many existing methods based on static FCs modeled by PCCs and based on dynamic FCs (the HMMs and LSTM).
A DGM with temporal dynamics will be explored in future work~\cite{Krishnan2017}.

\subsection{Reconstruction of Signals and Contributing Regions}\label{sec:reconstruction}
As shown in Fig.~\ref{fig:timeseries}, the reconstructed time-series were apparently similar to the originals, regardless of the class label.
The reconstruction error $\mathcal E(i,t,k;y)$ of the left anterior cingulum (which had the lowest contribution weight) was insensitive to the given class label.
In contrast, the reconstruction error $\mathcal E(i,t,k;y)$ of the left thalamus was certainly smaller with the correct label $y=1$ than with the incorrect label $y=0$.
The proposed DGM found a small but clear difference in the ROI-wise features between patients and controls despite the training procedure, in which the proposed DGM was not trained to discriminate these two entities.
We visualized the contribution weight $\mathcal W(i,t,k)$ at each frame $t$ in the bottom panels of Fig.~\ref{fig:timeseries}.
The contribution weight is always almost zero in a less contributing region (left anterior cingulum) in the right panels.
This means that the reconstruction of this region is independent from the class label $y$.
In the case of a largely contributing region (the left thalamus) in the left panels, the contribution weight $W(i,t,k)$ varies over time.
The contribution weight $W(i,t,k)$ is almost zero at $t=30$--$32$ and $t=34$--$37$, while it becomes very large at several frames such as $t=33$ and $38$.
We assumed that each frame is associated with the class label $y_i$ individually in Fig.~\ref{fig:gmodel} and Eq.~\eqref{eq:scanlikelihood}.
The objective in Eq.~\eqref{eq:wholelikelihood} is not discriminative (i.e., to classify each frame) but generative (i.e., to represent each frame).
Hence, if no disorder-related signals is found in a frame $x_{i,t}$, our proposed DGM can simply ignore the class label $y_i$ to represent the frame $x_{i,t}$.
Our proposed DGM successfully captured the temporally varying relationship between the class label $y_i$ and the frames $x_{i,t}$ (even though it did not capture the temporal dynamics).

As summarized in Fig.~\ref{fig:contribution_schizophrenia}, the proposed DGM found that the signals obtained from the thalamus ($\mathrm{Thalamus\_L}$ and $\mathrm{Thalamus\_R}$) significantly contributed to the correct diagnoses of schizophrenia.
This result agrees with many previous studies that have demonstrated the relationship between schizophrenia and the thalamus~\cite{Brickman2004,Glahn2008}.
While other regions are less significant than thalamus, they have been also mentioned in previous studies, e.g., fusiform in \cite{Quintana2003}, and temporal gyrus~\cite{Rajarethinam2000,Onitsuka2004}.
On the other hand, according to Fig.~\ref{fig:contribution_schizophrenia}, many regions are related to bipolar disorder at a similar level, but no regions are much more significant than others.
These regions have been mentioned in previous studies, e.g., cerebellum~\cite{Johnson2015,Laidi2015}, frontal inferior gyrus~\cite{Roberts2017}, and thalamus~\cite{Radenbach2010}.
Cerebellum is mainly related to sensorimotor system, but recent studies suggested that it is also related to emotion and behavior (and psychiatric disorders)~\cite{Turner2007,Barch2014}.
Therefore, we conclude that the proposed DGM identified the brain regions related to each disorder and we hope that these results encourage further biological investigations.

Several studies have already employed DNNs for diagnosis of neurological and psychiatric disorders and have attempted to identify regions and activity related to disorders.
Suk et al.~\cite{Suk2016} identified the contributing regions according to the weight parameters of the AEs: If several units representing brain regions in the input layer were connected to the same unit in the first hidden layer via large weight parameters, the regions were considered to contribute to the diagnosis with large weights.
However, if such a hidden unit had another largely biased input or a large bias parameter, the unit would be saturated after the activation function and could not transfer meaningful information to the subsequent layer, i.e., the unit would be ``dead'' (see Chapter 6, \cite{Gibson2017}).
A hidden unit also does not function when it is connected to the next hidden unit via a near-zero weight parameter.
Unlike PCA, the DNNs extract nonlinear and higher-order features not only in the first hidden layer but also in the subsequent layers.
The units and layers where features are extracted and whether the extracted features are actually used in the following layers are essentially uncontrolled~\cite{Zhao2017b,Ghorbani2017}.
As such, one cannot quantitatively compare contribution weights between multiple input units.
Conversely, the proposed DGM used region-wise reconstruction errors for the diagnosis.
Hence, the regions with large reconstruction errors certainly contributed to the diagnosis and the reconstruction errors corresponded to the contribution weights of the regions.

\begin{table*}[t]\centering
  \caption{Diagnostic Results of Non-balanced Measures.}\label{tab:appendresults}
  \setlength{\tabcolsep}{3pt}
  \begin{tabular}{lccccccccccccc}
  \toprule
    \multirow{2}{*}{\textbf{Model}}& \multicolumn{5}{c}{\textbf{Schizophrenia}} && \multicolumn{5}{c}{\textbf{Bipolar Disorder}}\\
    \cmidrule(lr){2-6}\cmidrule(lr){8-12}
      &\textbf{ACC} & \textbf{SPEC} & \textbf{SEN} & \textbf{PPV} & \textbf{NPV} &
      &\textbf{ACC} & \textbf{SPEC} & \textbf{SEN} & \textbf{PPV} & \textbf{NPV}\\
    \midrule
      chance level
      & 0.709 & ---   & ---   & ---   & --- &
      & 0.718 & ---   & ---   & ---   & --- \\
    \midrule
      PCC+Kendall+LLE+c-means~\cite{Shen2010}
      & 0.628 & 0.583 & 0.738 & 0.427 & 0.850&
      & 0.566 & 0.496 & 0.748 & 0.371 & 0.840\\
      PCC+SCCA+SLR~\cite{Yahata2016}
      & 0.720 & 0.799 & 0.529 & 0.535 & 0.810&
      & 0.687 & 0.802 & 0.396 & 0.445 & 0.776\\
    \midrule
      MLP
      & 0.753 & 0.978 & 0.202 & 0.534 & 0.754&
      & 0.732 & 0.992 & 0.069 & 0.230 & 0.732\\
      GMM
      & 0.675 & 0.773 & 0.438 & 0.465 & 0.775&
      & 0.730 & 0.995 & 0.052 & 0.225 & 0.729\\
      AE+MLP
      & 0.778 & 0.917 & 0.442 & 0.696 & 0.807&
      & 0.721 & 0.915 & 0.227 & 0.357 & 0.759\\
    \midrule
      HMM~\cite{Suk2016}
      & 0.600 & 0.629 & 0.530 & 0.376 & 0.767&
      & 0.720 & 0.939 & 0.158 & 0.372 & 0.742\\
      HMM(FO)+SVM~\cite{Vidaurre2017}
      & 0.712 & 0.823 & 0.441 & 0.511 & 0.786&
      & 0.609 & 0.663 & 0.474 & 0.390 & 0.757\\
      LSTM~\cite{Dvornek2017}
      & 0.725 & 0.810 & 0.518 & 0.574 & 0.808&
      & 0.675 & 0.791 & 0.380 & 0.446 & 0.766\\
    \midrule
      DGM (proposed)
      & 0.766 & 0.849 & 0.585 & 0.619 & 0.838&
      & 0.631 & 0.621 & 0.659 & 0.443 & 0.819\\
    \bottomrule
  \end{tabular}
\end{table*}

\subsection{Potential and Limitation of Proposed DGM}
Our proposed DGM is not robust to correlated regions since it assumes a diagonal covariance matrix for the posterior distribution $p_\theta(x_{i,t}|z_{i,t},y)$.
For example, when the left thalamus is parcellated into two regions, each of them has a similar influence on the diagnosis.
Hence, when directly applying our proposed DGM to voxel-wise features or 3D images, a large region could have a large contribution to the diagnosis compared to a small region, providing misleading results.
DGMs with non-zero covariances or implicit distributions could overcome this issue~ \cite{Goodfellow2014,Tran2017,Dorta2018}.
DGMs with temporal dynamics could also relax the assumption that the prior $p(z)$ of the frame-wise variabilities $z$ is time-invariant.

As mentioned in Section~\ref{sec:preprocessing}, our proposed DGM can accept states of dynamic FCs.
This could provide interpretable results such as FCs related to the disorder.
However, many existing methods to extract dynamic FCs are based on predefined models such as sliding window and HMM~\cite{Damaraju2014,Li2015f,Sourty2016,Vidaurre2017,Zhang2017m}, which could limit the flexibility of subsequent data-driven analysis by DNNs.
A dynamic FC extraction based on DNNs is a potential future work.

The current structure of our proposed DGM is designed for resting-state fMRI data and is not suited for a task-related fMRI dataset because it cannot accept the information about the imposed tasks.
When applying our proposed DGM to a task-related fMRI dataset without modification, task-related signals could be assigned with the frame-wise variability.
On the other hand, our proposed DGM can be easily extended to the task-related fMRI data by adding a new node $a_{i,t}$ representing the task imposed to a subject $i$ at time $t$.
Similarly to the class label $y_i$, the task information $a_{i,t}$ is a known factor that the input $x_{i,t}$ is assigned with while it can vary over time.

The proposed DGM can be trained as a discriminative model: Such a learning procedure increases the risk of overfitting, but potentially improves the classification accuracy by discriminating class labels~\cite{Lasserre2006,Prasad2017,Matsubara2017StockPred}.

\section{Conclusion}
This study proposed a deep neural generative model (DGM) for diagnosing psychiatric disorders.
The DGM was a generative model implemented using deep neural networks.
The DGM modeled the joint distribution of rs-fMRI images, class labels, and remaining frame-wise variabilities.
Using Bayes' rule, the DGM diagnosed test subjects with higher accuracy than other competitive approaches.
In addition, the DGM visualized brain regions that contribute to accurate diagnoses, which motivates further biological investigations.

\section*{Acknowledgments}
The authors would like to thank Dr.~Ben Seymour, Dr.~Kenji Leibnitz, Dr.~Hiroaki Mano, and Dr.~Ferdinand Peper at CiNet for valuable discussions.
The authors would appreciate the thoughtful comments of the anonymous reviewers.
This study was partially supported by the JSPS KAKENHI (16K12487) and the MIC/SCOPE \#172107101.

The authors declare that the research was conducted in the absence of any commercial or financial relationships that could be construed as a potential conflict of interest.

\appendices
\section{Detailed Results}\label{sec:appendix}
In this Appendix, we summarize the results of non-balanced measures in Table~\ref{tab:appendresults}.
Note that these scores show some tendencies of the models but are inappropriate for performance evaluations because the datasets are highly imbalanced.
For performance evaluations, see Tables~\ref{tab:results_schizophrenia} and \ref{tab:results_bipolar}.



\end{document}